\documentclass[review]{elsarticle}

\usepackage{lineno,hyperref}
\modulolinenumbers[5]

\usepackage{times}
\usepackage{epsfig}
\usepackage{graphicx}
\usepackage{amsmath}
\usepackage{amssymb}
\usepackage{color,soul}
\graphicspath{fig/}
\usepackage{subcaption}
\usepackage{booktabs}
\DeclareMathOperator*{\argmax}{argmax} 
\usepackage{multirow}

\journal{Journal of Pattern Recognition}

\begin{document}

\begin{frontmatter}

\title{Human Object Interaction Detection using Two-Direction Spatial Enhancement and Exclusive Object Prior}

\author[mymainaddress]{Lu Liu\corref{mycorrespondingauthor}}
\cortext[mycorrespondingauthor]{Corresponding author}
\ead{lliu@u.nus.edu}

\author[mymainaddress,mysecondaryaddress]{Robby T. Tan}
\ead{robby.tan@yale-nus.edu.sg}

\address[mymainaddress]{Electrical and Computer Engineering Department, National University of Singapore\\
E4-06-20, 4 Engineering Drive 3, 117583, Singapore}
\address[mysecondaryaddress]{Yale-NUS College, Singapore}

\begin{abstract}
Human-Object Interaction (HOI) detection aims to detect visual relations between human and objects in images. One significant problem of HOI detection is that non-interactive human-object pair can be easily mis-grouped and misclassified as an action, especially when humans are close and performing similar actions in the scene. To address the mis-grouping problem, we propose a spatial enhancement approach to enforce fine-level spatial constraints in two directions from human body parts to the object center, and from object parts to the human center. At inference, we propose a human-object regrouping approach by considering the object-exclusive property of an action, where the target object should not be shared by more than one human. By suppressing non-interactive pairs, our approach can decrease the false positives. Experiments on V-COCO and HICO-DET datasets demonstrate our approach is more robust compared to the existing methods under the presence of multiple humans and objects in the scene.

\end{abstract}


\end{frontmatter}


\section{Introduction}
Understanding relationships between individual objects is one of the important and challenging problems in visual recognition. To understand what is happening in a scene, computers need to recognize how humans interact with surrounding objects. In this paper, we tackle the challenging task of detecting human-object interactions (HOI), which is to understand the interactions between humans and objects. Given an image, HOI detection aims to detect an interaction triplet $<human, action, object>$. This requires to not only localize a human and an object instance, but also recognize the actions/interactions that the human is performing to the object, such as "ride bike" and "eat apple". HOI detection is an important step towards the high-level semantic understanding of human-centric scenes. 

HOI detection problem is challenging since it requires to correctly detect and recognize fine-grained human-centric interactions (e.g. "feed horse", "straddle horse" and "jump horse"), and also involves multiple co-occurring actions (e.g. "reading book while sitting on a chair"). Moreover, multiple humans can appear in one scene performing the same interaction to the same object ("throw and catch ball"), or performing the same interaction to the different objects ("people ski in the field"). These complex and diverse interaction scenarios impose significant challenges to HOI detection task.

\begin{figure}[ht!] 
         \centering 
         \includegraphics[width=1\columnwidth]{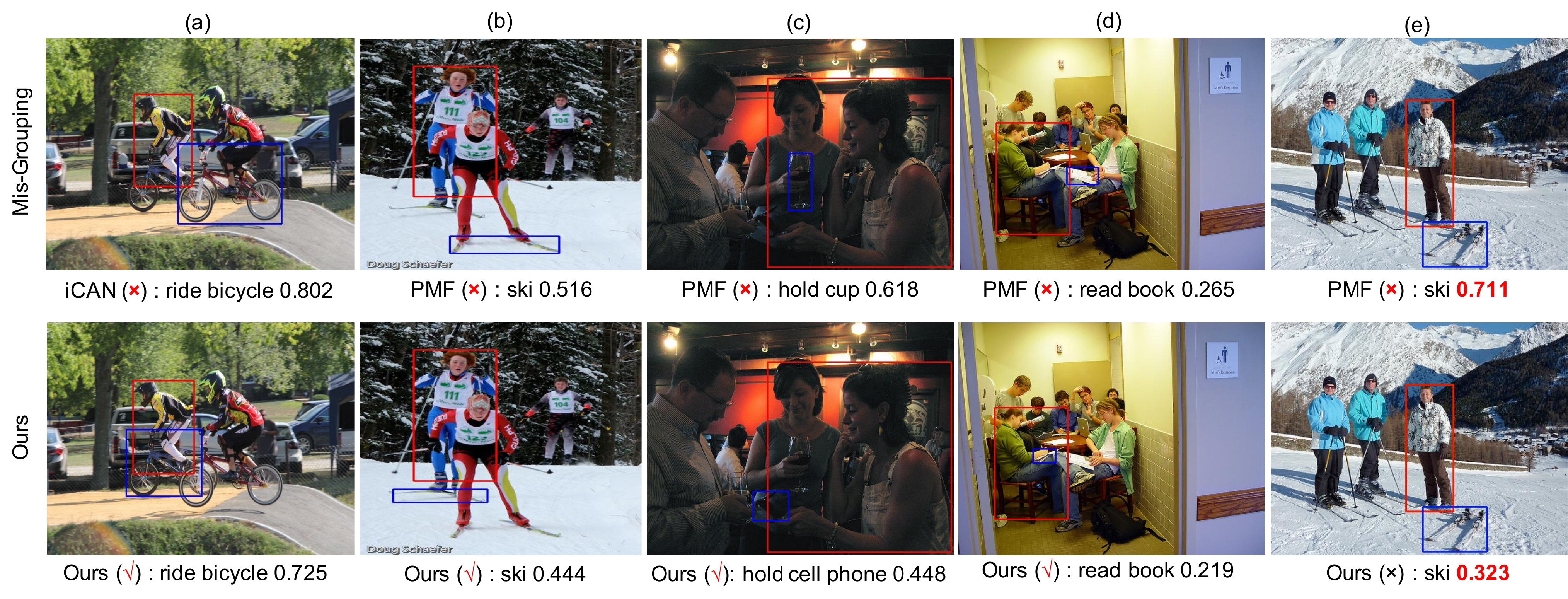}
         \caption{Example of HOI detections using the proposed approach, and iCAN~\cite{gao2018ican}, and the recently introduced PMFNet method~\cite{wan2019pose}. The first row shows five typical mis-grouping mistakes made by iCAN~\cite{gao2018ican} and PMFNet~\cite{wan2019pose}. Existing methods fails to correctly pair the agent human (red box) and the target object (blue box). The mis-grouping can happen when the wrong target object is close to the interested human (a-e), or the correct target object is occluded (a-d). Our approach accurately groups the human and the respective object (row 2 a-d), and suppresses the score for the non-interactive pair even when no positive target object exists in the scene (row 2 e). Better viewed in color.}
        \label{fig:eg} 
\end{figure}

One of the significant problems in HOI detection is mis-grouping (see Fig.~\ref{fig:eg}), which is no interaction human-object pair can be easily mis-grouped and misclassified as an action. The mis-grouping is one of the main causes of False Positives (FP) in HOI detections. We consider the mis-grouping error mainly comes from the complexity nature of combination, where multiple humans can interact with multiple objects via multiple actions/interactions. Specifically, when multiple objects lie around one human, this human can interact with any of these objects at the same time. When multiple humans and objects simultaneously appear in one scene, it becomes even harder to tell which human is performing which action to which object. 

Given a human of interest (red box in Fig.~\ref{fig:eg}) and an action of interest (e.g. ski), we want to find to which object (blue box in Fig.~\ref{fig:eg}) this human is performing this action. The paring correctness depends on the correctness of the HOI scoring function, which is to predict a higher score for the correct human-object pair than the wrong ones. However, when multiple humans appear in the scene, existing HOI detection methods~\cite{gao2018ican,wan2019pose,gkioxari2018detecting,xu2019learning,wang2019deep} can fail to identify the correct interaction pair, especially when humans are close and performing similar actions (e.g. people skiing in the field, or studying in a room in Fig.~\ref{fig:eg}). We consider the spatial affinity, object occlusion and appearance similarity are the main causes that can weaken the confidence score of the correct human-object pair, resulting in a false positive HOI detection result (see Fig.~\ref{fig:eg} row 1).

\footnotetext[1]{The score shown in Fig~\ref{fig:eg} (row 1 a) of iCAN has been normalized from $[0,2]$ to $[0,1]$ for comparison.}

To address the mis-grouping problem, InteractNet~\cite{gkioxari2018detecting} propose max-object selection to only select one target object, which is the highest scoring object, for each human-action pair to form the final HOI triplet. Intuitively, for example, one human cannot ride more than one bike at the same time. It assumes that there is no such cases as human walking multiple dogs or herding multiple sheep. This max-object selection can filter out the mis-grouped detections made in iCAN~\cite{gao2018ican} shown in Fig.~\ref{fig:eg} (a)\footnotemark[1]. Although this max-object selection can suppress non-interactive target objects for each human, it cannot suppress non-interactive pairs between one object and multiple humans. For instance, two humans skiing one pair of skis dose not comply with the common sense.
TIN~\cite{li2019transferable} suppresses the mis-grouped non-interactive pairs by training a binary interactiveness classifier to first filters out non-interactive pairs predicted by the interactiveness network using a threshold, and then predict HOI actions for the predicted interactive pairs. However, this first-stage hard filtering can exclude true positive pairs as well, and the interactiveness threshold setting is hand-crafted and empirical. The recent PMFNet~\cite{wan2019pose} jointly trains the binary interactiveness classifier with the HOI action classifier, and fuse the scores as the final prediction. However, when multiple humans are performing similar actions in one scene, PMFNet~\cite{wan2019pose} fails to correctly pair the agent human and the target object. For instance, in Fig.~\ref{fig:eg} (b), two humans are skiing together. PMFNet~\cite{wan2019pose} fails to pick out the correct background skis for the background human, because the background skis are partially occluded, and the foreground skis are close to the background human. Moreover, when the human is not interacting with any objects in the scene such as in Fig.~\ref{fig:eg} (d), it becomes even harder to suppress such non-interactive pair, since the scene is just like skiing, and the skis are close to the human feet.

In this paper, our goal is to suppress non-interactive human-object pairs to address mis-grouping problem. We propose two approaches during training and inference respectively to not only predict a small action score for non-interactive human-object pairs, but also pick out the correct interactive target object for the interested human. 

Firstly, we propose a spatial enhancement approach to enforce fine-level spatial constraints in two directions from human body parts to the object center (i.e. center point of the object box), and from object parts to the human center. Spatial relationship is an important cue to infer HOI actions. For example, if a human is riding a bike, the human should be just above the bike, and the central axis of human and object should be close to each other (Fig.~\ref{fig:eg} (a)). Previous methods~\cite{gao2018ican,xu2019learning,chao2018learning,li2019transferable} only encode the rough spatial locations between the whole human bounding box and object bounding box. Recent methods extract spatial features between human body parts and the object center~\cite{wan2019pose,zhou2019relation}. However, we argue that purely encoding the object center is not distinguishing enough when the object is large and surrounding the human, such as "lie in bed" and "drive car". Also for objects that have several separate components, such as "a pair of skis", the center point cannot describe the shape and the exact location of the object instance. Hence, we propose symmetric spatial constraints from human parts to the object center, and from object parts to the human center to enhance the spatial configurations between the target human and object. By encoding the detailed part-level spatial features, the network can learn more discriminative spatial features, which can help predict high confidence scores for interactive human-object pairs and low scores for non-interactive pairs.

To further suppress non-interactive pairs, we propose a human-object regrouping approach by considering the prior knowledge of the exclusive object of an action. For instance, as shown in Fig.~\ref{fig:eg} (b), existing method PMFNet~\cite{wan2019pose} predicts both humans in foreground and background are skiing the foreground skis. However, this detection result does not comply with the common sense, because one pair of skis cannot be ridden by more than one human at the same time. Motivated by this observation, our idea is to take the object exclusion property into consideration. When predicting an object-exclusive action (e.g. "ski", and "read book"), the prior knowledge tells us the object should be exclusive to one human, and cannot be shared by multiple humans at the same time. Hence, if there are multiple humans sharing the same object instance, we assign this object to the max-scoring human, and reselect the next scoring target object for the other non-max humans. 

Compared to previous methods~\cite{chao2018learning,gkioxari2018detecting,gao2018ican,xu2019learning,chao2018learning,li2019transferable,wan2019pose,zhou2019relation,wang2019deep}, our approach enhances the spatial configurations between human and object, and takes advantage of action-specific exclusive object prior knowledge. Our approach can correct the mis-grouping errors (Fig.~\ref{fig:eg} row 1) made in existing methods~\cite{gao2018ican,wan2019pose} and accurately groups the human and the corresponding object under the presence of multiple humans and objects with occlusions in the scene (Fig.~\ref{fig:eg} row 2). Hence, our approach can decrease the false positives in HOI detection results. 

Our main contributions are three fold: (1) To address the mis-grouping problem, we propose a two-direction spatial enhancement approach to enforce fine-level spatial constraints between human keypoints and object parts. (2) we propose an action-specific exclusive object prior knowledge to regroup the object with the corresponding interactive human. (3) Extensive experiments on V-COCO~\cite{gupta2015visual} and HICO-DET~\cite{chao2015hico} datasets demonstrate our approach is more robust compared to the existing methods under the presence of multiple humans and objects in the scene.

\section{Related work}
\label{sec:related_work}
{\bf Object detection.} Object detection~\cite{girshick2015fast,ren2015faster,dai2016r,girshick2014rich,lin2017feature} is an essential building block for scene understanding. Our work uses the off-the-shelf Mask RCNN~\cite{he2017mask,wu2019detectron2} to detect humans and object instances. Given the detected instances, our method aims to recognize interactions between all pairs of human and object instances.

{\bf Image Human Action Recognition.} There has been great progress in understanding human actions from images in recent years~\cite{gkioxari2015actions,zhao2017single,liu2019loss}. Compared to action recognition from videos~\cite{tran2015learning,carreira2017quo}, which highly relies on motion, action recognition from a single image depends on static cues, such as human poses~\cite{girdhar2017AttentionalPoolingAction}, body parts~\cite{gkioxari2015actions,zhao2017single}, and contexts~\cite{gkioxari2015contextual}. R*CNN~\cite{gkioxari2015contextual} discover possible interactive objects around the human as context cues to infer human actions. Zhao et al.~\cite{zhao2017single} incorporate mid-level body part actions to infer body actions. Liu et al.~\cite{liu2019loss} address the misleading object problem by jointly predicting human action and location, to drive the attentions of the network to the target human. Compared to general image action recognition, HOI detection also utilizes the similar visual cues, such as human poses, body parts, and contexts. Differently, HOI further requires to detect individual instances, and identify the corresponding interactive objects in the human-centric scene.

{\bf Visual Relationship Detection.} Visual Relation Detection (VRD) aims to detect pairs of objects and classify their relationships simultaneously. Lu et al.~\cite{lu2016visual} propose to learn language priors from semantic word embedding to refine relationship detection. Vtrans~\cite{zhang2017visual} has a visual translation network to model relation as a translation vector from subject to object. ~\cite{zhan2019exploring} exploits undetermined relationships, including unlabeled pairs, irrelevant pairs, and falsely detected object pairs, to help predict visual relationships. Recently, ~\cite{zhang2019large} leverages language models to learn separate visual-semantic embeddings for subject, object and predicate. Peyre et al.~\cite{peyre2019detecting} propose to learn an additional embedding space for visual phrase classes. They further develop a model using analogy embedding transfer to infer unseen relationships. Different from general relationships detection between arbitrary objects, in our task, we focus on human-centric relationship detection, which aims to infer how humans interact with objects.

{\bf Human-Object Interaction Detection.} Gupta and Malik~\cite{gupta2015visual} first explore the problem of visual semantic role labelling and construct V-COCO dataset for the role labelling task. In this problem, the goal is to detect the agent (human), an object, and predict the interaction class between them. Later, Chao et al.~\cite{chao2018learning} introduce a new benchmark dataset HICO-DET, and propose a multi-stream network to aggregate human, object and spatial configurations for HOI detection. Gkioxari et al.~\cite{gkioxari2018detecting} propose InteractNet to estimate action-specific density map over target locations based on detected human appearance. Some works~\cite{qi2018learning,xu2019learning,zhou2019relation} model the scene as a graph reasoning problem and use graph convolutional network to predict the HOI. ~\cite{shen2018scaling,kato2018compositional} adopt compositional learning strategy for zero-shot learning to predict unseen human-object combinations. Context-based methods mine the contextual information in the image to infer HOI. iCAN~\cite{gao2018ican} designs an instance-centric attention module to highlight important regions in the image as contextual information for facilitating interaction prediction. DCA~\cite{wang2019deep} extracts context-aware appearance features using squeeze-and-excitation operation. Part-based approaches~\cite{wan2019pose,fang2018pairwise,zhou2019relation} utilize human poses to extract detailed human body part features. Our work is, in particular, related to PMFNet~\cite{wan2019pose} exploiting human keypoints to extract detailed part features to predict actions. However, in contrast to PMFNet~\cite{wan2019pose}, we additionally encode the spatial relationship from object parts to the human center to better recognize the fine-grained interactions and suppress the non-interactive pairs. TIN~\cite{li2019transferable} shares the same the goal as ours, which is to suppress non-interactive human-object pairs. However, TIN~\cite{li2019transferable} requires the availability of multiple datasets to train the interactiveness network, and has a transfer stage to fine-tune on the target dataset. In contrast, our approach is simple and effective, which only learns interactiveness knowledge on the target dataset, and the interactiveness network is jointly trained with action classification network.


\section{Method}
\label{sec:method}

\begin{figure}[t!] 
         \centering 
         \includegraphics[width=1\columnwidth]{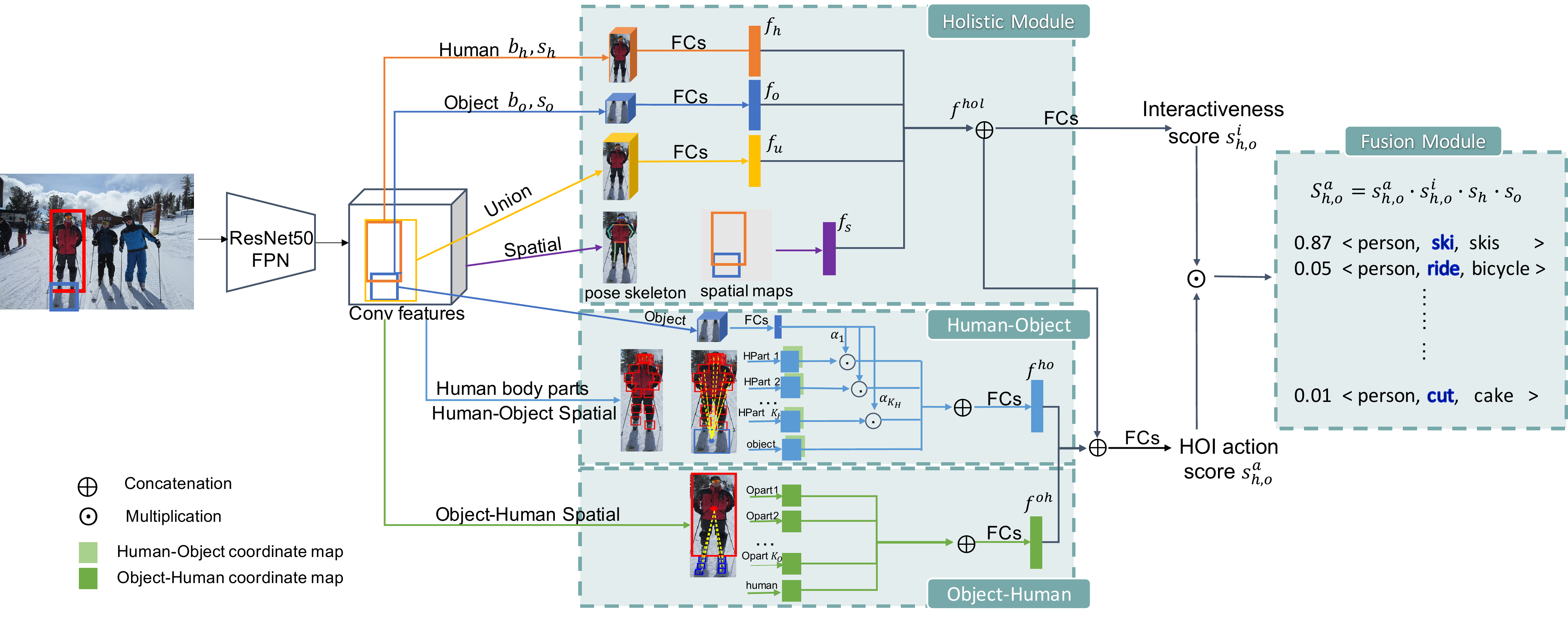}
         \caption{Overview of our framework. Given a pair of human-object proposal, holistic module employs human, object, union, and spatial-pose branches to extract appearance features from human, object and union boxes, and spatial features from pairwise location masks and human pose information. We employ two-direction spatial constraints symmetrically from human parts to the object center, and from object parts to the human center. We jointly predict interactiveness score and HOI action score. Finally, we combine the interactiveness score and action score to predict the final score for HOI actions. Better viewed in color.}
        \label{fig:network} 
\end{figure}

We now introduce our relation reasoning method for human-object interaction detection. The task is to detect humans and objects in the input image, as well as to recognize the actions for each $<human, object>$ pair. We develop a multi-branch deep neural network (Fig.~\ref{fig:network}) leveraging multi-level features with holistic instance-level appearance features, human part-level appearance features, and spatial features in both instance and part levels. We present the overview of our method in Sec.~\ref{sec:overview}.

To address the mis-grouping problem, we propose a novel symmetric two-direction spatial enhancement module to encode the fine-level spatial configurations between human parts and object parts. We explain our spatial encoding method in details in Sec.~\ref{sec:spatial}. Furthermore, in the inference stage, we exploit the exclusive object prior knowledge to regroup human and object instances to further suppress the non-interactive human-object pair. We describe our regrouping strategy in Sec.~\ref{sec:regroup}.

\subsection{Overview}
\label{sec:overview}
We develop a bottom-up framework in which we first detect all the object proposals, and then predict the interactions between each $<human, object>$ pair. Given an input image, we first employ Mask R-CNN from Detectron~\cite{wu2019detectron2} to detect all the human/object bounding boxes $b_h, b_o$, human keypoints $K_{p_h}$, and object instance segmentation masks $m_o$. We use $s_h$ and $s_o$ to denote the confidence scores for a detected human box $b_h$ and object box $b_o$, respectively. All the human and object proposals will be paired up to form $<human, object>$ pairs as HOI proposals. In the interaction classification stage, for each human-object bounding box pair $(b_h, b_o)$, we predict the action score $s^{a}_{h,o}$ for each HOI action category $a$. Besides, we jointly predict a binary interactiveness score $s^{i}_{h,o}$ indicating whether the human is interacting with the object or not. The final HOI score $S^{a}_{h,o}$ is the combination of the confidence scores of the proposals $s_h, s_o$, interaction score $s^{a}_{h,o}$, and the interactiveness score $s^{i}_{h,o}$:
\begin{equation} \label{eq:score}
S^{a}_{h,o} = s^{a}_{h,o} \cdot s^{i}_{h,o} \cdot s_h \cdot s_o
\end{equation}

\subsection{Holistic Visual Representations}
We adopt ResNet-50-FPN~\cite{lin2017feature} as our convolutional backbone network to generate convolution feature maps $\Phi$ with dimension of $(H \times W \times D)$. We use Mask R-CNN~\cite{he2017mask} as the object detector to generate human-object pairs ${(b_h, b_o)}$. Given the detected human/object bounding boxes $b_h, b_o$, the holistic module consists of four branches: human branch, object branch, union branch, and spatial branch, as illustrated in Fig.~\ref{fig:network}. We extract the appearance features of human proposal $b_h$, object proposal $b_o$, and union proposal $b_u$ from the shared feature map of ResNet-50 Conv5 by the ROI-Align operation~\cite{he2017mask} with resolution of $D_h \times D_h$. The union proposal $b_u$ is the minimum box in the spatial region that contains $b_h$ and $b_o$, which can the extract contextual information around humans and objects in the scene.

In addition to appearance features, spatial features can also provide important cues for recognizing interactions, such as "ride bike" vs. "walk bike". As for non-interactive pairs, if a human is far from a bike, he can hardly ride or walk the bike. And, if a human is riding a bike, the bike should be just below the human body, and not in front or behind the human body. We apply the two-channel binary masks human and object proposals in their union space to capture the spatial configurations of each human-object instance as in~\cite{chao2018learning,gao2018ican,wan2019pose}. In addition, the pose structure of the human can also help differentiate interactions, such as "hit with the baseball bat" vs. "look at the tennis bat". As for non-interactive pairs, if a human is squatting down on the ground, he can hardly hitting with the bat. Following~\cite{li2019transferable,wan2019pose}, we represent the estimated human poses with $K_H=17$ keypoints as a skeleton line-graph. The intensity values range from 0.05 to 0.95 in a uniform interval indicating different human parts. Finally, the binary masks and pose skeleton map are rescaled to 64x64 and concatenated channel-wise to generate holistic spatial features.

For each branch in the holistic module, we employ two fully connected layers (FC) to embed the features into a low-dimensional space. We denote the output features of human, object, union, spatial branch as $f_h, f_o, f_u, f_s$. All the features are concatenated ($\oplus$) to form the final holistic feature $f^{hol}=f_h \oplus f_o \oplus f_u \oplus f_s$.

\subsection{Two-Direction Spatial Enhancement}
\label{sec:spatial}
\begin{figure} [t!]
         \centering 
         \includegraphics[width=0.8\columnwidth]{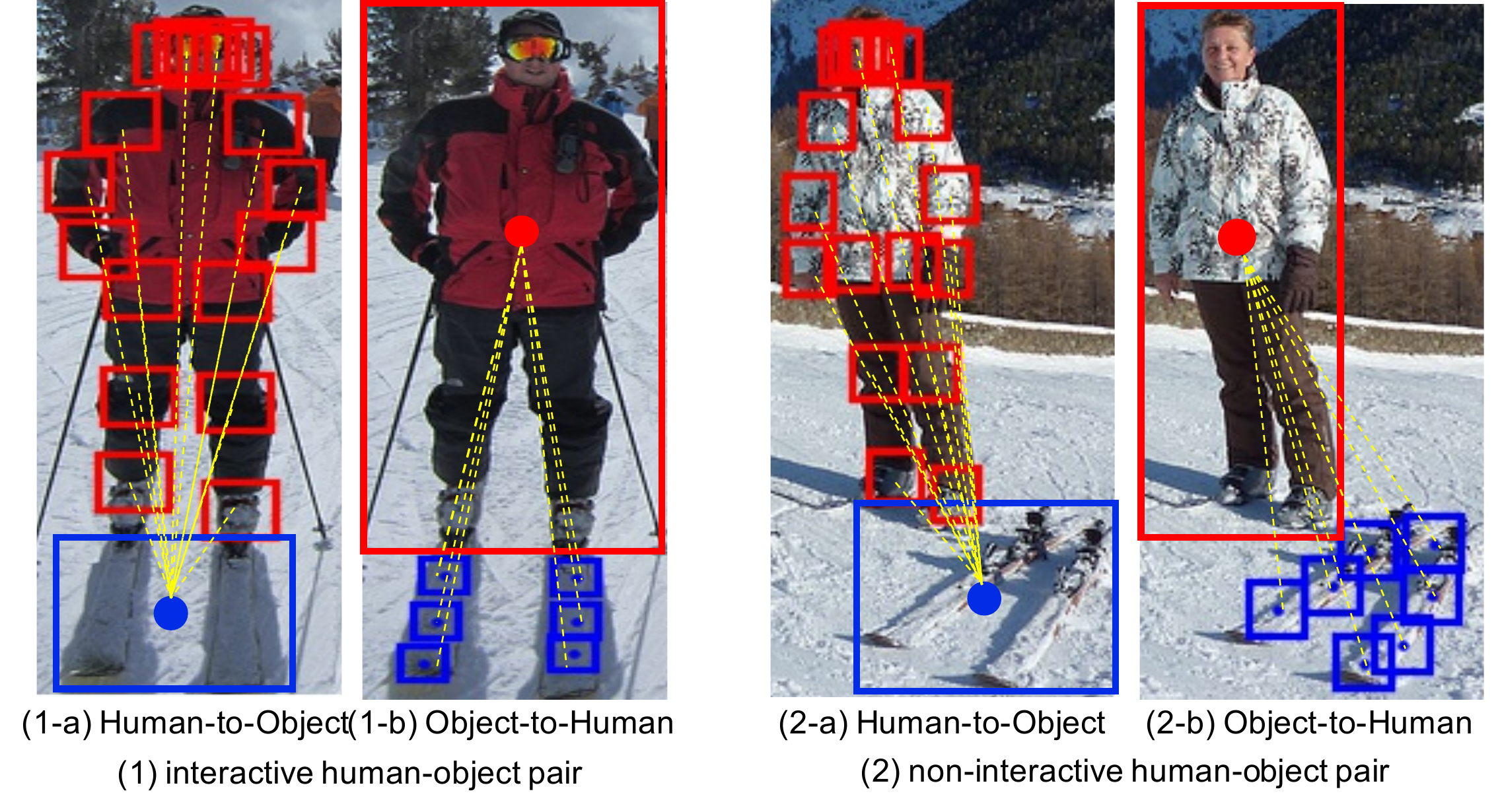}
         \caption{Our proposed symmetric two-direction spatial enhancement can describe the detailed spatial distributions between human body parts and object parts. We show an interactive human-object example in (1). The pair of skis are close to the human feet, and the two ski boards are located on the two sides of the human body. However, in the non-interactive example (2), only using human part to the object center cannot tell whether the human is standing on the ski board or not, because the object center in (2-a) is as close as in (1-a) to the human feet. We additionally encode the relative spatial locations from object parts to the human center shown in (2-b), which can tell that the two ski boards only lie on one side of the human. Hence, (2) is likely to be no-interaction.}
        \label{fig:spatial} 
\end{figure}

Our goal is to suppress non-interactive human-object pairs. To this end, our idea is not only to encode the rough spatial configurations between the human box and object box, but also to model the detailed interaction patterns between the human parts and object parts. Recent methods~\cite{wan2019pose,zhou2019relation} extract spatial features from human body parts to the object center. However, we argue that purely encoding the center point is not distinguishable enough when the object is large and surrounding the human, such as "lie in bed" and "drive car". Also for objects that have several separate components, such as "a pair of skis", the center point cannot describe the shape of the object instance. For example, "human ski" shown in Fig.~\ref{fig:spatial}, the spatial relationships between human parts to the object center is not distinct enough to differentiate between the interactive pair (in Fig.~\ref{fig:spatial} (1-a)) and non-interactive pair (in Fig.~\ref{fig:spatial} (2-a)), because the object center in both cases are close to the human feet. 

Our method enhances the spatial interaction patterns in two directions: from human keypoints to the object center, and from the object parts to the human center. By comparing Fig.~\ref{fig:spatial} (1-b) and (2-b), we can see that the spatial distributions from object parts to the human center can explicitly tell the network that when the two ski boards lie on one side of the standing human, the human is probably not skiing with them.

\begin{figure} 
         \centering 
         \includegraphics[width=0.8\columnwidth]{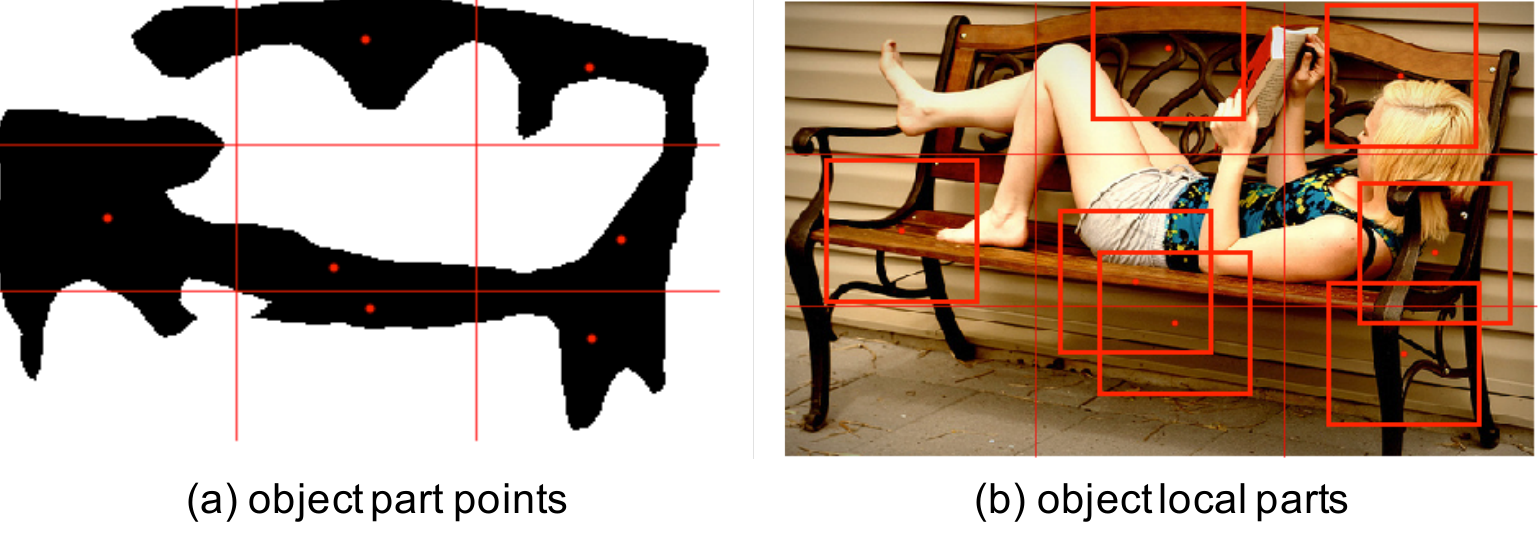}
         \caption{An illustration of our object parts on the bench instance using 3x3 grid ($R=3$). The red points are valid object part points, and red boxes in (b) are the valid object parts with 0.1 object box size. Better viewed in color.}
        \label{fig:parts} 
\end{figure}

We first generate object parts from the segmentation mask of the object $b_o$. The segmentation is a list of polygon vertices around the object as in the COCO format~\cite{lin2014microsoft}. Given a $b_o$ and its segmentation polygons, we first convert the segmentation from the polygon format to a binary segmentation mask $m_o$ with the same size as $b_o$. If a pixel $i$ is within the object region, $m_o(i)=1$, else $m_o(i)=0$. Then, we use a regular 2D grid size of $R \times R$ to separate the mask $m_o$ into $K_O = R^2$ bins. For each bin, we use the mean point of the mask inside the bin as our object part points $o_k$. Along with the part points, we generate a flag (length of $R^2$) for each part indicating the validness of the part. There are two requirements for a valid object part $o_k$ where $flag[o_k]=1$: (1) the mean point is inside the object region, i.e. $m_o(o_k)=1$, (2) and the number of object pixels inside the bin is larger than a minimum percentage $r$ of the bin size $\sum_i^{i \in bin} m_o(i) > r R^2$. We further define a local box region $b_{o_k}$ for each valid part point $o_k$, which is a box centered at $o_k$ with a size $\gamma$ proportional to the size of the object proposal $b_o$. An illustration of object part points and local parts is shown in Fig.~\ref{fig:parts}. Similarly, we generate human local part boxes by defining local boxes $b_{h_1},..., b_{h_{K_H}}$ around all the human keypoints, as shown in Fig.~\ref{fig:spatial} (1-a) and (2-a).

With the generated human and object parts, we can now encode the spatial relationships between them. We first generate an object-center-normalized coordinate map for the human-object branch, and a human-center-normalized coordinate map for the object-human branch. Then, we obtain the spatial offset maps for the human parts and object parts by cropping part boxes from the two coordinate maps, respectively. 

In particular, for each input human-object box pair $(b_h, b_o)$, supposing the center point location of $b_h$ and $b_o$ is $(x_h, y_h)$ and $(x_o, y_o)$, respectively, we generate a two-channel coordinate map $Coord = (Coord_x, Coord_y)$ representing the $x, y$ coordinates for each pixel in $\Phi$. Each map has the same spatial dimension $H \times W$ as the convolutional feature map $\Phi$. Then, we compute the relative spatial offset maps in two directions: from the human local parts to the object center $Coord^{ho} = (Coord^{ho}_x, Coord^{ho}_y)$, and from the object local parts to the human center $Coord^{oh} = (Coord^{oh}_x, Coord^{oh}_y)$.
\begin{equation} \label{eq:spatial}
\begin{split}
Coord^{ho}_x = Coord_x - x_o, &\quad\quad\quad Coord^{ho}_y = Coord_y - y_o, \\
Coord^{oh}_x = Coord_x - x_h, &\quad\quad\quad Coord^{oh}_y = Coord_y - y_h.
\end{split}
\end{equation}

Then, in human to object direction, we obtain the spatial offset map $Coord_{h_k}^{ho}$ for human part $h_k$ and $Coord_o^{ho}$ for object by cropping (ROI-Align~\cite{he2017mask}) each human part $b_{h_k}$ and object box $b_o$ from the object-center-normalized coordinate map $(Coord^{ho}_x, Coord^{ho}_y)$, with the same spatial dimension $D_p \times D_p$. In parallel, in object to human direction, we generate spatial offset map $Coord_{o_k}^{oh}$ for object part $o_k$ and $Coord_h^{oh}$ for human by cropping each object part region $b_{o_k}$ and human box $b_h$ from the human-center-normalized map $(Coord^{oh}_x, Coord^{oh}_y)$.

We distribute the two direction spatial configurations into two branches. In human-object branch, we adopt ROI-Align for the created human part boxes $b_{h_1}, ..., b_{h_{K_H}}$ together with object box $b_o$ on the convolution feature map $\Phi$ to obtain appearance features $\{f^{app}_{h_1},..., f^{app}_{h_{K_H}}\}$ and $f^{app}_{h_o}$ where each feature is of size $D_p \times D_p \times D$. Then, for each human part $h_k$, we obtain human part features $f_{h_k}$ by concatenating the part appearance feature $f^{app}_{h_k}$ with human to object spatial feature $Coord^{ho}_{h_k}$. Instead of treating all the human body parts equally, we predict the human body part attention guided by the object appearance. Our idea is that different object usually interact with different human parts. For instance, given a "book", if we want to predict "read book", we need to look at human hands and eyes to decide. Hence, object can infer the importance of different human parts. We predict human part attention using a small network consists of two FC layers with sigmoid function to normalize the attention values to $[0,1]$. The predicted attention value for part $h_k$ is denoted as $\alpha_k$. The final feature $f_{h_k}$ for each human part $h_k$ contains both part appearance feature and spatial offsets, as in Eq.~\ref{eq:human}. Finally, we concatenate all human part features with object feature, and feed it into several FC layers to extract final local feature $f^{ho}$. 

In object-human branch, to enhance spatial feature from object parts to human, we concatenate the spatial offsets for all object parts and feed into several FC layers to extract final local feature $f^{oh}$. Note, for invalid object parts, we remain both appearance and spatial features to be zero, so that the network parameters will not be trained using those invalid parts.
\begin{equation} \label{eq:human}
\begin{split}
f_{h_k} &= \alpha_k \cdot (f^{app}_{h_k} \oplus Coord^{ho}_{h_k}) \\
f_{h_o} &= f^{app}_{h_o} \oplus Coord_o^{ho} \\
f^{ho} &= FC(f_{h_1} \oplus f_{h_2} ... \oplus f_{h_{K_H}} \oplus f_{h_o}) \\
f^{oh} &= FC(f_{o_1} \oplus f_{o_2} ... \oplus f_{o_{K_O}} \oplus Coord^{oh}_h)
\end{split}
\end{equation}

Finally, we take the holistic features $f_{hol}$ and feed it into the interactiveness classification head (consists of two FC layers followed by sigmoid function $\sigma$) to generate the interactiveness score $s^i_{h,o}$. We predict HOI action score $s^a_{h,o}$ from all relation features, including holistic $f^{hol}$ and part-level features $f^{ho}$ and $f^{oh}$.
\begin{equation} \label{eq:pred}
\begin{split}
s^i_{h,o} &= \sigma(FC(f^{hol})) \\
s^a_{h,o} &= \sigma(FC(f^{hol} \oplus f^{ho} \oplus f^{oh}))
\end{split}
\end{equation}

\subsection{Model Training}
In the training stage, the training data consist of ground-truth HOI annotations, labeled as $<human, action, object>$ triplets, where the humans and objects are localized as bounding boxes. Besides the ground-truth instance bounding boxes, we also augment the HOI training samples using the object proposals given by the detector. We freeze ResNet-50 backbone network, and only train our HOI classification network to predict HOI action categories. Note that the object detector head, pose estimation head, and instance segmentation head in Mask R-CNN~\cite{he2017mask} are external modules and are not involved in the training process.

Assume for a training sample with a form of human-object pair $(b_h,b_o)$, the ground-truth action label is an action set $a_{GT}=\{a_1,...,a_n\}$, where $n$ is the number of ground-truth actions. We form the target HOI action label vector as $\mathbf{y}=[y^1, ...,y^A]$, where $y^a = 1$ for $a \in a_{GT}$, and $A$ is the total number of HOI aciton categories. We also generates the binary interactiveness label $z \in {0,1}$. We define $z=1$ if $a_{GT} \neq \emptyset$, else $z=0$.
 
We adopt holistic features $f_{hol}$ to infer the interactiveness score, and both holistic $f_{hol}$ and part-level features $f^{ho}$ and $f^{oh}$ for HOI action score. Suppose the predicted HOI action score for HOI action category $a$ is $s^a_{h,o}$, and interactiveness score is $s^i_{h,o}$. Our network jointly optimizes HOI action classification loss $L_{cls}$, and binary interactiveness loss $L_{inter}$. The HOI classification is a multi-label classification problem, since a human can perform multiple actions to one or multiple target objects simultaneously, e.g. a person can "hit with" and "hold" a baseball bat at the same time. We apply a binary sigmoid classifier for each action category, and then minimize the cross-entropy loss between action score and the ground-truth action label for each action category as:
\begin{equation} \label{eq:loss}
L = \sum^A_{a=1} L_{cls} (y^a, s^a_{h,o}) + \lambda L_{inter} (z, s^i_{h,o}),
\end{equation}
where both $L_{cls}$ and $L_{inter}$ are cross-entropy loss computed as $L_{CE} (a,b) = a \log(b) + (1-a) \log(1-b)$, and $\lambda$ is the loss weight to balance the contribution of the multi-label interaction prediction and binary interactiveness prediction.

\subsection{Exclusive Object Regrouping}
\label{sec:regroup}

Our goal is to suppress the non-interactive pairs to address the mis-grouping problem. Besides enhancing the spatial interaction patterns during training, we also propose a human object regrouping method in the inference stage by considering the exclusion property of the object conditioned on the interested action. 

At inference, given an input image with $N_h$ human proposals, $N_o$ object proposals, we can form $N_h \times N_o$ human-object candidate pairs. For each pair $(b_h, b_o)$, the network predicts an action score $s^a_{h,o}$ for HOI action class $a$, corresponding to the $<human, action, object>$ HOI triplets using Eq.~\ref{eq:score}. Thus, for each image, we obtain a score matrix $S$ size of $N_h \times N_o \times A$, where each entry of $S$ is $S^a_{h,o}$, and $A$ is the total number of HOI action categories. Then, for each human-action pair $(h,a)$, we apply max-object selection~\cite{gkioxari2018detecting} to select the object box that maximizes $S^a_{h,o}$:
\begin{equation} \label{eq:max_object}
b_{o^*} = \argmax_{b_o} S^a_{h,o}
\end{equation}

Intuitively, for each human and each action, this max-object selection only remains the max-scoring object as the target object. For example, one human cannot ride more than one bike at the same time. It assumes that there is no cases like human is walking multiple dogs or herding multiple sheep. This max-object selection can reduce the potential triplet number from $N_h \cdot N_o \cdot A$ to $N_h \cdot A$.

\begin{figure} [t!]
         \centering 
         \includegraphics[width=1\columnwidth]{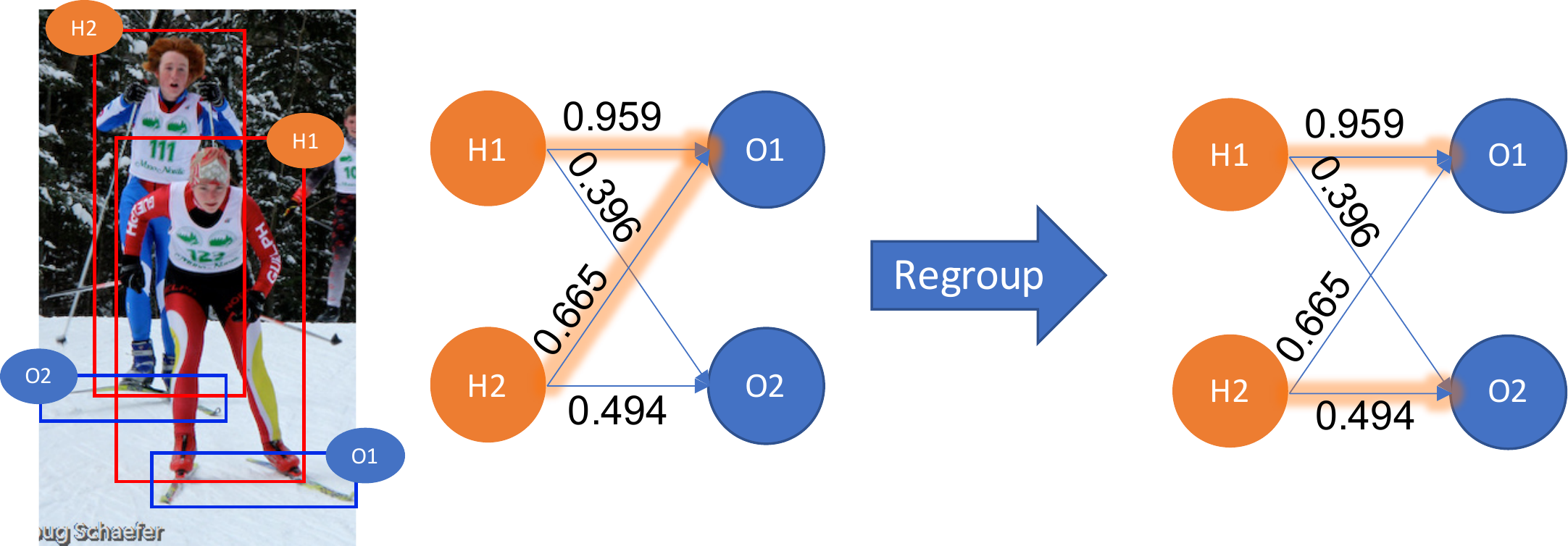}
         \caption{Our proposed exclusive object regrouping approach. The score on the edge is the predicted score of the human-object pair for action "ski". The highlighted edge is the predicted human-object pair for action "ski". Since "ski" is an object-exclusive action, i.e. one pair of skis cannot be shared by more than one human, our method regroup the mis-grouped non-interactive pair between $H2$ and $O1$, and correctly pair the background human $H2$ with her skis $O2$ in the background. Better viewed in color.}
        \label{fig:regroup} 
\end{figure}

We further introduce our novel exclusive object regrouping algorithm by considering the exclusion property of the object. As illustrated in Fig.~\ref{fig:regroup}, common sense tells us that one pair of skis cannot be ridden by more than one human at the same time. In order to correct the hard false positive pair between $H2$ and $O1$, our idea is to reselect the next scoring target object for the other non-max humans in the image.

We first obtain the exclusive object prior knowledge from the statistics of the dataset. From the HOI ground-truth annotations, we compute for each action $a$, how many humans share the same object instance in the image. We compute two statistics in the whole dataset: (1) exclusive object number: the number of ground-truth human-object pairs where one object is only paired with one human, denoted as $Q^a_e$; (2) sharable object number: the number of ground-truth pairs where one object is paired with more than one human, denoted as $Q^a_s$. If for action $a$, the majority of human-object pairs are exclusive, defined as $Q^a_e / (Q^a_e+Q^a_s) > \beta$, then the action $a$ is regarded as an object-exclusive action. 

With the acquired object-exclusive action list, when predicting an object-exclusive action (e.g. ski, skaterboard), if there are multiple humans sharing the same object instance $b_{o^*}$ ($O1$ in Fig.~\ref{fig:regroup}), we assign this shared object $b_{o^*}$ to the max-scoring human $b_{h^*}$ ($H1$ in Fig.~\ref{fig:regroup}), and reselect the next scoring target object $b'_{o^*}$ ($O2$ in Fig.~\ref{fig:regroup}) for the other non-max humans ($H2$ in Fig.~\ref{fig:regroup}) in the image in a descending order. Thus, basically, we find the one-to-one matching between human and object instances for object-exclusive actions. To avoid wrong replacement for the non-max humans, we only apply the regrouping algorithm when the prediction score of the max-scoring human-object pair is confident enough, i.e. $S^a_{b_{h^*}, b_{o^*}}> S_{min}$.

\section{Experiments}
\label{sec:exp}

{\bf Dataset.} We evaluate our method on V-COCO~\cite{gupta2015visual} and HICO-DET~\cite{chao2015hico} datasets. \textbf{V-COCO dataset} (Verbs in COCO), which is a subset of COCO~\cite{lin2014microsoft}, includes 5,400 train+val images and 4,946 test images. It is annotated with 26 unique verb classes, and has bounding boxes for humans and interacting objects. In particular, three verb classes (i.e. cut, hit, eat) are annotated with two types of targets (i.e. instrument "instr" and direct object "obj"). \textbf{HICO-DET dataset}~\cite{chao2015hico} consists of 47,776 images with more than 150K human-object pairs (38,118 images in training set and 9,658 in test set). It is annotated 600 HOI categories over 80 object categories (same as in COCO~\cite{lin2014microsoft}) and 117 unique verbs. The bounding boxes of humans and corresponding objects are also annotated.

{\bf Evaluation Metrics.} We follow the standard evaluation metric as in~\cite{gupta2015visual,gkioxari2018detecting} and report mean Average Precision (mAP) to measure the HOI detection performance. The mAP is computed based on both recall and precision, which is appropriate for the detection task. A HOI triplet $<human,action,object>$ is considered as a true positive only when both the human and object detected bounding boxes have IoUs (intersection-over-union) greater than 0.5 with the respective ground-truth and the associated interaction class is correctly classified.

\subsection{Implementation Details}
We use Mask R-CNN~\cite{he2017mask} from Detectron~\cite{wu2019detectron2} with a feature backbone of ResNet-50-FPN~\cite{lin2017feature} to detect object proposals, human keypoints, and the object instance segmentation mask. We implement our network based on ResNet-50-FPN~\cite{lin2017feature} to be comparable to ~\cite{gkioxari2018detecting,wan2019pose}. We crop the ROI features from the highest resolution feature map in FPN with channel dimension $D=256$. We set the ROI-Align resolution in the holistic module to be $D_h=7$, and in human-object and object-human part-level modules, we set resolution to be $D_p=5$. The dimension of features $f_{h}, f_o, f_u, f_s, f^{oh}$ is set to 256, and $f^{ho}$ is set to 512. The size of the local part boxes for human/object is $\gamma=0.1$ of human/object box length=max(height, width). We use $ 3 \times 3$ grid (i.e. $R=3$) to generate object parts, and minimum ratio for a valid bin $r=1/16$. During testing, the human and object bounding boxes with detection confidence scores above 0.5 and 0.4 respectively are kept. 

In VCOCO, out of 26 HOI action categories, we obtain 13 object-exclusive actions from annotation statistics, including "drink-instr", "eat-instr", "hit-instr", "hit-obj", "hold-obj", "jump-instr", "read-obj", "skateboard-instr", "ski-instr", "snowboard-instr",  "talk on phone-instr", "throw-obj", "work on computer-instr". In HICO-DET, out of 600 HOI categories, we obtain 372 object-exclusive actions, such as "read/open book", "hold keyboard/wine glass", and "drive car/bus". We set the minimum confidence score between max human-object pair as $S_{min}=0.5$, and the statistic ratio of an object-exclusive action as $\beta=95\%$.

We freeze ResNet-50-FPN backbone, and train the parameters of the HOI classification network. In VCOCO, we use SGD optimizer for training with initial learning rate 4e-2, weight decay 1e-4, and momentum 0.9. We reduce the learning rate to 4e-3 at iteration 24k, and stop training at iteration 48k. In HICO-DET, we use SGD optimizer for training with initial learning rate 5e-2, weight decay 1e-4, and momentum 0.9. We reduce the learning rate to 5e-3 at iteration 30k, and stop training at iteration 90k. The ratio of positive and negative (i.e. non-interactive) samples is 1:3. The weight in the loss function is set as $\lambda=0.1$. 

\subsection{Quantitative Results}

\begin{table}[!t]
\centering
\caption {Results comparisons on V-COCO~\cite{gupta2015visual} and HICO-DET~\cite{chao2015hico} datasets. We report mAP$_{role}$ on V-COCO, and mAP on HICO-DET for the full HOI categories in both default setting (object unknown) and Know Object setting. Higher values indicates better performance.}
\label{tab:results}
\begin{tabular}{lc|cc}
\toprule
&V-COCO& \multicolumn{2}{c}{HICO-DET} \\
Methods &  mAP$_{role}$ & \quad Default \quad & Known Object  \\ \midrule \midrule
VRSL~\cite{gupta2015visual} (impl. by~\cite{gkioxari2018detecting}) & 31.8 &--&--\\
Shen et al.~\cite{shen2018scaling} & -- & 6.46&-- \\
HO-RCNN~\cite{chao2018learning} & -- &7.81& 10.41 \\
InteractNet~\cite{gkioxari2018detecting} & 40.0 &9.94&--\\
BAR~\cite{kolesnikov2019detecting} & 41.1 &--&--\\
GPNN~\cite{qi2018learning} & 44.0 &13.11&--\\
iCAN~\cite{gao2018ican} & 45.3 &14.84&16.26\\
Xu et al. ~\cite{xu2019learning} & 45.9 &14.70&--\\
PMFNet-Base~\cite{wan2019pose} &48.6&14.92&18.83 \\
DCA~\cite{wang2019deep} & 47.3 &16.24&17.73\\
RPNN ~\cite{zhou2019relation} & 47.5 &17.35&--\\
No-Frills~\cite{gupta2019no} &-- &17.18 &--\\
TIN ($RP_DC_D$) ~\cite{li2019transferable} & 47.8 &17.03&19.17\\ 
PMFNet ~\cite{wan2019pose} & 52.0 &17.46&20.34\\ \midrule
Ours & \textbf{52.28} &\textbf{17.55}&\textbf{20.61}\\
\bottomrule
\end{tabular}
\end{table}

We first compare our proposed approach with state-of-the-art methods in literature. Tab.~\ref{tab:results} shows the comparison on the V-COCO and HICO-DET datasets. On V-COCO, among existing approaches, VRSL~\cite{gupta2015visual} uses spatial constraints for the interacting objects. BAR~\cite{kolesnikov2019detecting} applies box attention based on Faster R-CNN framework. iCAN~\cite{gao2018ican} utilizes human, object appearance and pairwise spatial configuration with contextual attention. The RPNN approach~\cite{zhou2019relation} using attention graphs for parsing relations of object and human body-parts obtains a mAP$_{role}$ of 47.5. TIN~\cite{li2019transferable} transfers the interactiveness network trained from multiple datasets to the target dataset to perform non-interaction suppression, and reports a mAP$_{role}$ of 47.8. The recent PMFNet~\cite{wan2019pose} utilizes human poses to extract body part appearance and keypoints to object spatial feature, and achieves mAP$_{role}$ of 52.0. Our method achieves superior performance compared to PMFNet~\cite{wan2019pose} with a mAP$_{role}$ of 52.28, which is the state-of-the-art result. 

On HICO-DET, we report the mAP results for the full HOI category with two different settings of Default and Known Objects. We observe that our proposed model achieves competitive results over the state-of-the-art approaches~\cite{zhou2019relation,wan2019pose,gupta2019no}. Our approach obtains mAP of 17.55, and 20.61 in Default and Known Object settings with absolute gains of 0.09 and 0.27 over PMFNet~\cite{wan2019pose}, respectively. 

\subsection{Ablation Study}
We perform several experiments to evaluate the effectiveness of our model components on V-COCO dataset (Tab.~\ref{tab:vcoco}). The baseline model consists of holistic module and human-object appearance feature, and achieves 51.67 mAP. Adding human-object spatial and object-human spatial improves absolute gain of 0.21 and 0.29, respectively. We further add both two-direction spatial branch, and achieves 52.18 mAP with 0.22 gain over baseline. Adding exclusive-prior regrouping, we obtain 52.16 mAP with 0.2 absolute gain compared to without regrouping (51.96). Using both our proposed spatial enhancement and exclusive prior regrouping, our full model achieves 52.28 mAP with 0.61 gain over baseline. These results demonstrate that our enhanced two-direction spatial configurations enables the network to suppress non-interactive human-object instances more effectively. By further considering the exclusive property of the object, our method can correct the mis-grouped pairs and associate the human to its associate target object.

\begin{table}[!t]
\centering
\caption {Ablation study of each component in our network on V-COCO dataset.}
\label{tab:vcoco}
\begin{tabular}{lcccccc}
\toprule
Add-on & baseline & \quad\quad & \quad\quad & \quad\quad & \quad\quad & \\ \midrule
holistic & $\checkmark$ & $\checkmark$ &$\checkmark$ & $\checkmark$&$\checkmark$&$\checkmark$ \\
human-object appearance & $\checkmark$ &$\checkmark$ & $\checkmark$&$\checkmark$&$\checkmark$& $\checkmark$ \\
human-object spatial & &&$\checkmark$ &$\checkmark$&$\checkmark$&$\checkmark$ \\
object-human spatial & &$\checkmark$ & &$\checkmark$ &&$\checkmark$ \\
exclusive-prior & &&&&\checkmark&\checkmark \\ \midrule
mAP$_{role}$  & 51.67 & 51.88 & 51.96 & \quad 52.18 &  \quad 52.16 & \quad 52.28 \\
\bottomrule
\end{tabular}
\end{table}

\subsection{Qualitative Visualizations}

\begin{figure}[t!]
         \centering 
         \includegraphics[width=1\columnwidth]{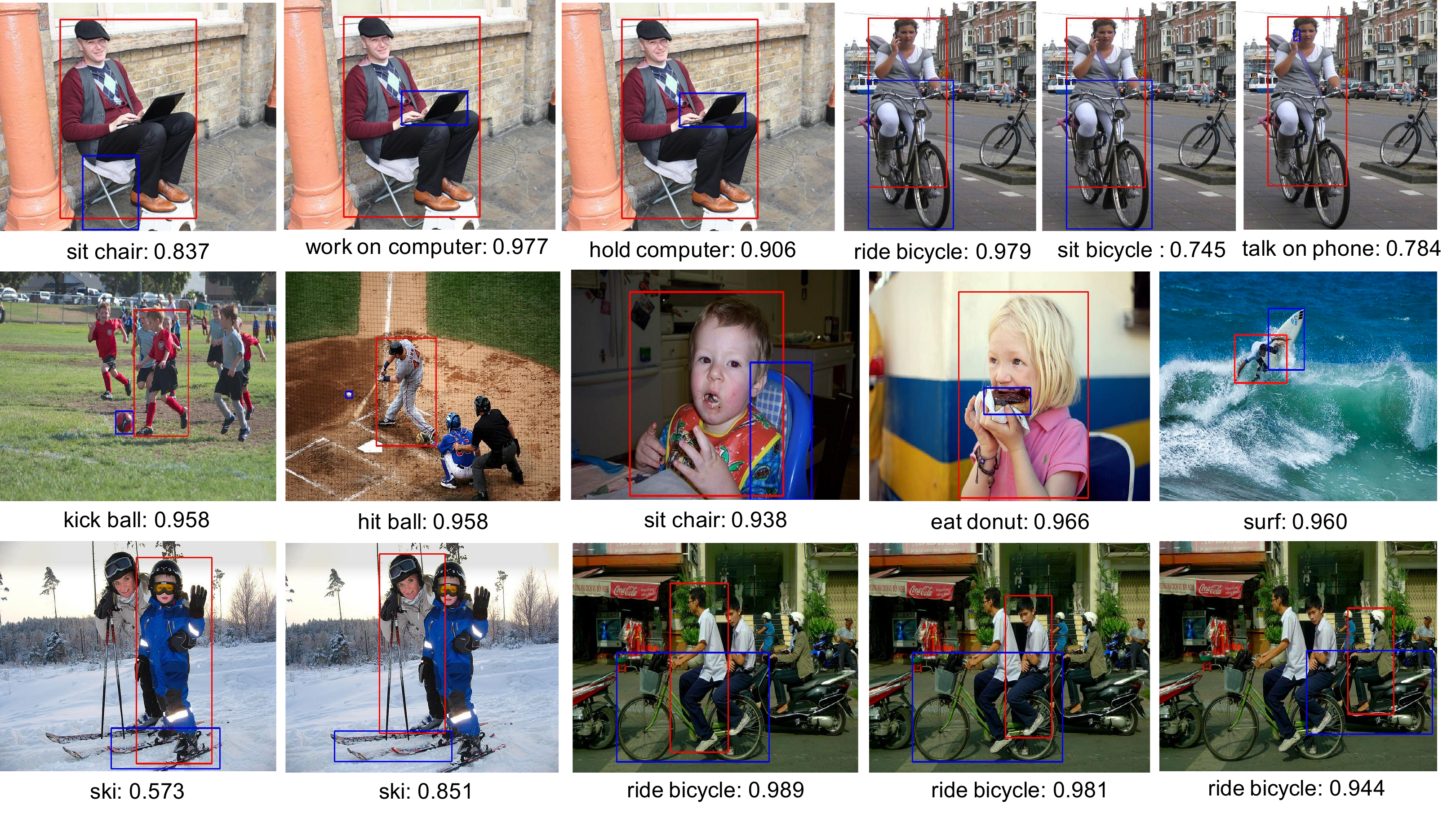}
         \caption{HOI detection results of our method on VCOCO. A human (red box) can perform multiple actions to multiple objects (blue box).  Our method can correctly pair the human with the corresponding target object when multiple humans in the scene. Better viewed in color.}
        \label{fig:exp_vcoco} 
\end{figure}

\begin{figure}[t!]
         \centering 
         \includegraphics[width=1\columnwidth]{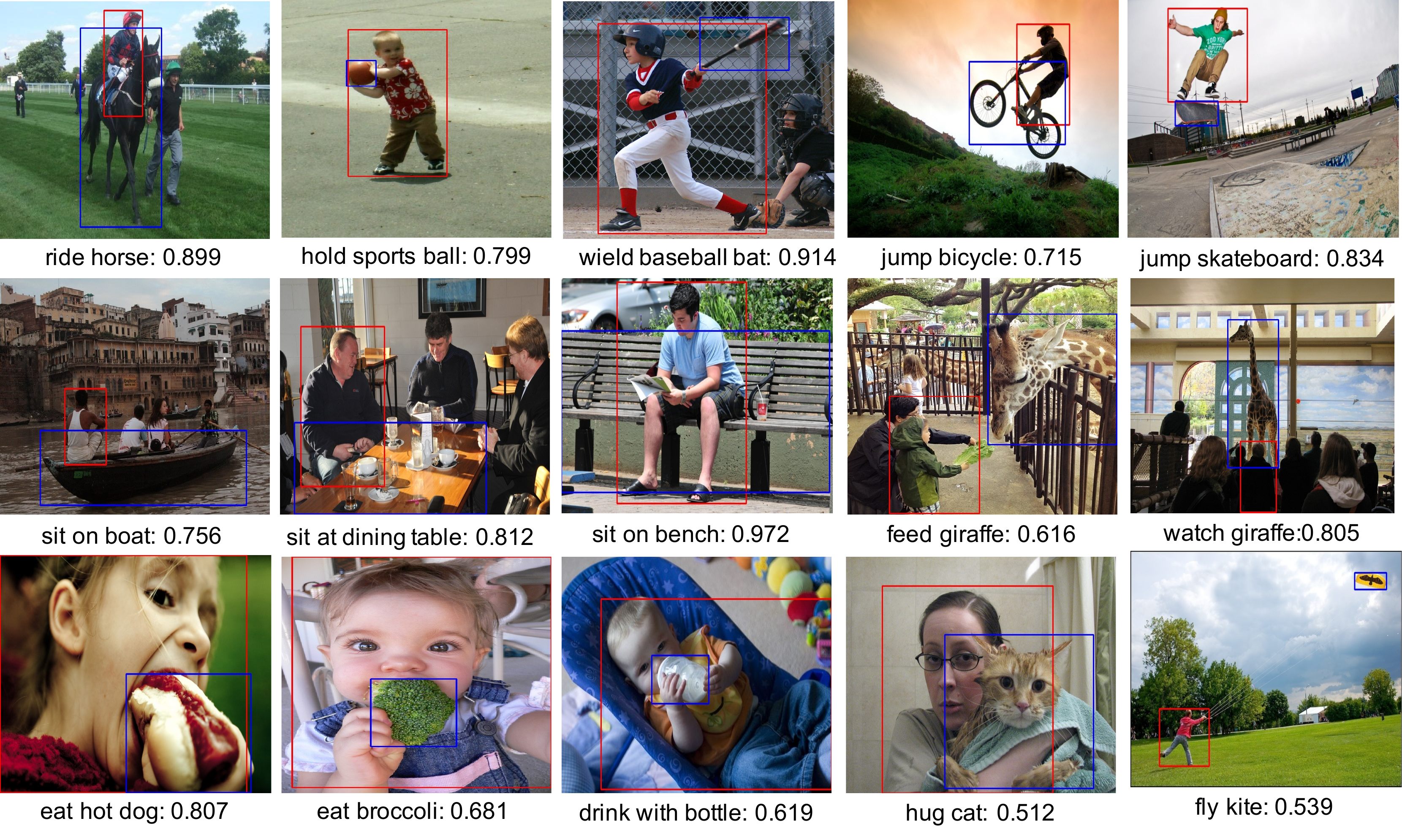}
         \caption{HOI detections results of our method on the HICO-DET. Our model can predict HOIs in a wide variety of different situations in daily life. Better viewed in color.}
        \label{fig:exp_hico_all} 
\end{figure}

We show our human-object interaction detection results on VCOCO in Fig.~\ref{fig:exp_vcoco}. Each subplot shows one detected $<human,action,object>$ triplet. The first row in Fig.~\ref{fig:exp_vcoco} shows our correctly detected triplets of one human taking multiple actions on multiple objects. When multiple humans appear in the scene, our method can correctly pair the interactive human-object pairs and recognize the action between them (Fig.~\ref{fig:exp_vcoco} row 3).

\begin{figure}[t!]
         \centering 
         \includegraphics[width=1\columnwidth]{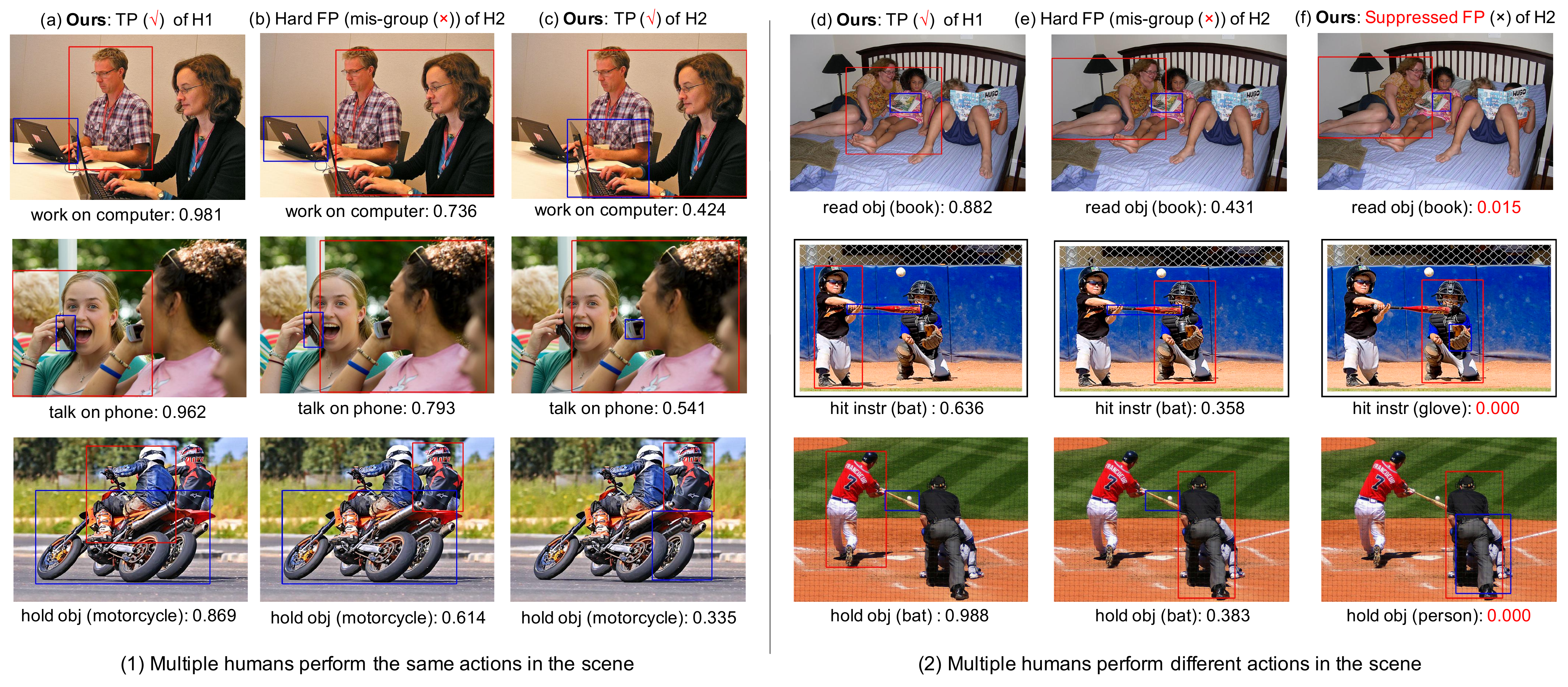}
         \caption{HOI detection results of our full model with and without exclusive object regrouping at inference on VCOCO. Our method can remove the mis-grouped false positive human-object pairs shown in column (b, e). In particular, on the left (1), we show three examples that multiple humans are performing the same actions in the scene. Our method can correctly associate the human (i.e. H2) with his true target object (see column (c)). One the right (2), we show three examples that multiple humans are performing different actions in the scene. In column (e), the human in red box is not performing the interested action, but can be mis-grouped with an object. For instance, in row 2, column (f), the child is not hitting anything. Our method successfully suppresses the non-interactive pairs with low confidence scores when the human is not performing the interested action (see column (f)). We can see that our proposed method can alleviate the mis-grouping issue in HOI detection. Better viewed in color.}
        \label{fig:exp_regroup} 
\end{figure}

\begin{figure}[t!]
         \centering 
         \includegraphics[width=1\columnwidth]{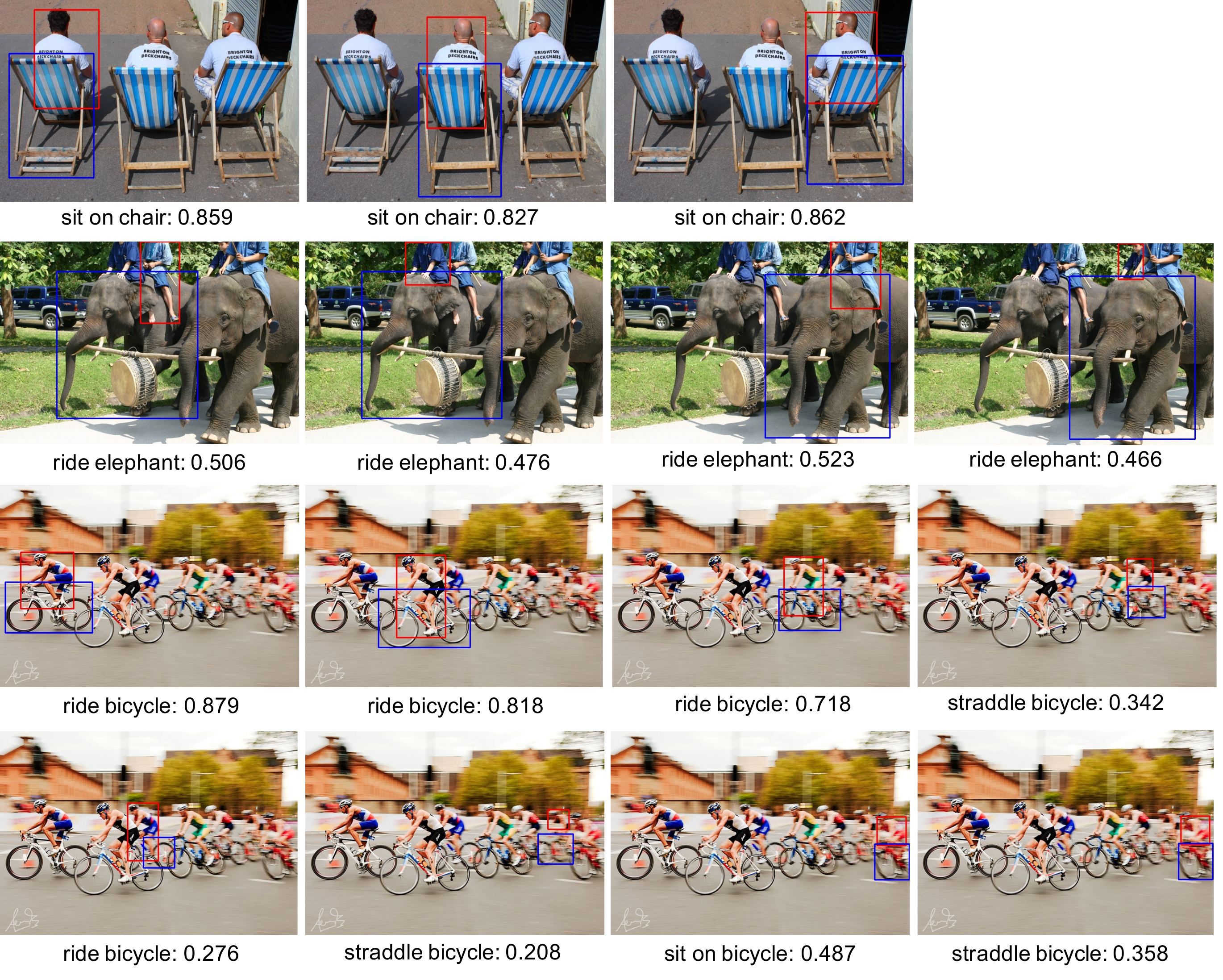}
         \caption{HOI detections results of our method on HICO-DET under the presence of multiple humans in the scene. We can see that our proposed method can correctly pair the interactive human-object pairs, even when they are performing similar actions with occlusions. Better viewed in color.}
        \label{fig:exp_hico} 
\end{figure}

Fig.~\ref{fig:exp_hico_all} shows our detection results on HICO-DET. It shows that our model can predict HOIs in a wide variety of different situations in daily life.

Fig.~\ref{fig:exp_regroup} shows the effectiveness of our proposed exclusive object regrouping method on VCOCO. Due to occlusion and spatial affinity between humans and objects, purely relying on the network prediction can mis-group non-interactive human and object with a high confidence, resulting in hard False Positive (FP) detections (see Fig.~\ref{fig:exp_regroup} (b, e)). Our method can correctly associate the human with its true target object at inference (Fig.~\ref{fig:exp_regroup} (c)), and suppress non-interactive pairs when the human is not performing the action interested (Fig.~\ref{fig:exp_regroup} (f)).

We further show more detection results under the presence of multiple humans in the scene on HICO-DET in Fig.~\ref{fig:exp_hico}. We can see that our proposed method can correctly pair the interactive human-object pairs under the presence of multiple humans in the scene, even when they are performing similar actions with occlusions.

\section{Conclusion}
\label{sec:conclusion}
In this paper, we address the problem of mis-grouping in human-object interaction detection task, where non-interactive pairs are wrongly paired and classified as an action. Our goal is to suppress non-interactive pairs, and hence can reduce false positive HOI detections. We propose  two-direction spatial constraints between human body parts and object parts. To further suppress the non-interactive pairs, at inference, we introduce an exclusive object regrouping approach by considering the exclusion property of the object conditioned on the interested action. Our approach is able to correct mis-grouped hard false positive human-object pairs when multiple humans performing similar actions in the scene, and suppress the non-interactive human-object pairs even when the target the object is close to the human. Experiments on V-COCO and HICO-DET datasets demonstrate our approach is more robust compared to the existing methods under the presence of multiple humans and objects in the scene.


\bibliography{mybib}

\end{document}